\documentclass[10pt,journal,transmag]{IEEEtran}
\pdfoutput=1
\usepackage[utf8]{inputenc}
\usepackage[T1]{fontenc}
\usepackage{amsmath}
\usepackage{amssymb}
\usepackage{amsfonts}
\usepackage{bm}
\usepackage{bbm}
\usepackage{epsfig}
\usepackage{grffile}
\usepackage{amsthm}
\usepackage[export]{adjustbox}

\usepackage{numprint}
\usepackage{enumitem}
\usepackage[noadjust]{cite}

\usepackage[usenames,dvipsnames]{color}
\definecolor{grey}{rgb}{.35,.35,.35}
\definecolor{dblue}{rgb}{0,0.1,.6}
\definecolor{dgreen}{rgb}{0,.6,0.1}

\usepackage[colorlinks=true,citecolor=dblue,linkcolor=dblue,urlcolor=dblue]{hyperref}
\usepackage[all]{hypcap}

\renewcommand{\vec}[1]{{\boldsymbol{#1}}}

\newcommand{\va} {\vec{a}}
\newcommand{\vf} {\vec{f}}
\newcommand{\vx} {\vec{x}}
\newcommand{\vphi} {\vec{\phi}}
\newcommand{\vvphi} {\vec{\varphi}}

\newcommand{\RR}{\mathbb{R}}
\newcommand{\CC}{\mathbb{C}}

\newcommand{\mc}[1]{\mathcal{#1}}

\newcommand  {\Pmatrix}[1]{\begin{pmatrix}#1\end{pmatrix}}

\makeatletter
\renewcommand{\p@subsection}{}
\renewcommand{\p@subsubsection}{}
\makeatother

\newcommand{\duke} {Department of Physics \& Duke Quantum Center,\\ Duke University,\\ Durham, North Carolina 27708, USA}

\newcommand{\ethz} {Department of Physics,\\ ETH Zurich,\\ 8093 Z\"urich, Switzerland}

\begin{document}

\title{Machine learning with tree tensor networks, CP rank constraints, and tensor dropout}
\author{\IEEEauthorblockN{Hao Chen}\IEEEauthorblockA{\ethz}
\and
\IEEEauthorblockN{Thomas Barthel}\IEEEauthorblockA{\duke}}

\date{February 15, 2024}

\maketitle

\begin{abstract}
\emph{Abstract}---Tensor networks developed in the context of condensed matter physics try to approximate order-$N$ tensors with a reduced number of degrees of freedom that is only polynomial in $N$ and arranged as a network of partially contracted smaller tensors. 
As we have recently demonstrated in the context of quantum many-body physics, computation costs can be further substantially reduced by imposing constraints on the canonical polyadic (CP) rank of the tensors in such networks [arXiv:2205.15296]. 
Here, we demonstrate how tree tensor networks (TTN) with CP rank constraints and tensor dropout can be used in machine learning. The approach is found to outperform other tensor-network-based methods in Fashion-MNIST image classification. A low-rank TTN classifier with branching ratio $b=4$ reaches a test set accuracy of 90.3\%  with low computation costs. Consisting of mostly linear elements, tensor network classifiers avoid the vanishing gradient problem of deep neural networks. The CP rank constraints have additional advantages: The number of parameters can be decreased and tuned more freely to control overfitting, improve generalization properties, and reduce computation costs. They allow us to employ trees with large branching ratios, substantially improving the representation power.
\end{abstract}

\section{Introduction}
Tensor networks were developed as a tool to describe states of strongly-correlated quantum many-body systems \cite{Orus2014-349}
in order to resolve the so-called curse of dimensionality. The latter refers to the fact that the dimension of the quantum state space grows exponentially with the number of particles. In many cases, the quantum states of interest can be approximated by a tensor network -- a network of partially contracted tensors, where the reduced number of parameters (the tensor elements) scales polynomially instead of exponentially in the system size. The expressiveness and computation costs of a tensor network method increase with increasing bond dimensions, which are the dimensions of vector spaces associated to the edges in the network.
Very accurate approximations can be achieved if the network structure is well-aligned with the entanglement structure of the studied system. Prominent classes of tensor networks are matrix product states (MPS) \cite{Baxter1968-9,Fannes1992-144,White1992-11,Rommer1997,PerezGarcia2007-7,Schollwoeck2011-326} a.k.a.\ tensor trains \cite{Oseledets2011-33}, tree tensor networks (TTN) \cite{Fannes1992-66,Otsuka1996-53,Shi2006-74,Hackbusch2009-15,Murg2010-82,Nakatani2013-138}, the multiscale entanglement renormalization ansatz (MERA) \cite{Vidal-2005-12,Vidal2006}, and projected entangled-pair states (PEPS) \cite{Niggemann1997-104,Nishino2000-575,Verstraete2004-7,Verstraete2006-96}.

The success of tensor network methods for the investigation of quantum matter motivated their recent adaptation to machine learning. First developments addressed both supervised learning \cite{Cohen2016-29,Stoudenmire2016-29,Novikov2016_05,Stoudenmire2018-3,Grant2018-4,Liu2019-21,Huggins2019-4,Efthymiou2019_06,Glasser2020-8,Selvan2020_04,Chen2020_11,Liu2021-7,Araz2021-2021,Cheng2021-103,Kong2021_01,Dborin2022-7,Convy2022-3a,Convy2022-3b,Dilip2022-4,Dymarsky2022-4,Strashko2022_08,Guala2023-13} and
unsupervised learning \cite{Han2018-8,Stoudenmire2018-3,Cheng2019-99,Sun2020-101,Bai2022-39,Shi2022-105,Fernandez2022-12,Vieijra2022_02,Lidiak2022_07,Liu2023-107,Shi2023_02} employing
MPS \cite{Stoudenmire2016-29,Novikov2016_05,Han2018-8,Stoudenmire2018-3,Huggins2019-4,Cheng2019-99,Efthymiou2019_06,Sun2020-101,Selvan2020_04,Chen2020_11,Liu2021-7,Araz2021-2021,Dborin2022-7,Bai2022-39,Convy2022-3a,Convy2022-3b,Shi2022-105,Fernandez2022-12,Dilip2022-4,Dymarsky2022-4,Lidiak2022_07,Strashko2022_08,Liu2023-107,Guala2023-13,Shi2023_02},
TTN \cite{Cohen2016-29,Stoudenmire2018-3,Grant2018-4,Liu2019-21,Huggins2019-4,Cheng2019-99,Convy2022-3b,Guala2023-13},
MERA \cite{Grant2018-4,Kong2021_01}, and
PEPS \cite{Cheng2021-103,Vieijra2022_02}.

In this work, we introduce and benchmark tensor network classifiers, where we impose constraints on the canonical polyadic (CP) rank of the tensors, i.e., use low-rank tensors\footnote{For an order-$z$ tensor $A\in\CC^{m_1}\otimes\dotsb\otimes\CC^{m_z}$, $(m_1,\dotsc,m_z)$ are the (bond) dimensions of $A$ and its CP rank $r$ is defined in Eq.~\eqref{eq:CPD}.}.
This is motivated by corresponding recent advances in tensor network state simulations of quantum many-body systems \cite{Chen2022_05}. The idea is borrowed from the canonical polyadic decomposition \cite{Hitchcock1927-6,Carroll1970-35,Harshman1970-16,Kolda2009-51}, which has a wide range of applications in data analysis \cite{Kolda2009-51}.
In particular, we work with TTN built from low-rank tensors. The image classifier maps each pixel into one component of an exponentially big tensor product space. Local low-rank tensors then reduce the number of components in each layer of the TTN by a factor of $1/b$. The output of the tensor in the top layer is the decision function.
The resulting low-rank TTN classifiers have several advantages:
\begin{enumerate}[leftmargin=3ex,itemsep=0em]
  \item
  TTN and MPS classifiers allow for an efficient evaluation of decision functions, with costs scaling linearly in the number $N$ of pixels for image classification. In contrast, the evaluation costs for PEPS and MERA scale exponentially in $N$ on classical computers unless one employs suitable approximations in the tensor contractions \footnote{MERA decision functions can be evaluated efficiently on quantum computers, and there exist approximate contraction methods for PEPS with polynomial costs \cite{Verstraete2004-7,Verstraete2006-96,Nishino1996-65,Orus2009_05}.}.
  \item
  MPS are primarily designed to encode correlations in one-dimensional (1D) systems and computation costs increase exponentially when MPS are applied for $D\geq 2$ dimensional systems. Similar problems will occur when applying MPS for machine learning on data that has $D\geq 2$ dimensional structure such as 2D images. TTN, MERA, and PEPS are more suitable to capture correlations/features for $D\geq 2$.
  \item
  For TTN, MPS, and MERA (but not PEPS), one can use the tensor-network gauge freedom such that all tensors become partial isometries; see, for example, Ref.~\cite{Barthel2022-112}. Hence, corresponding classifiers can be employed in quantum machine learning by implementing the isometric tensors with unitary gates \cite{Grant2018-4,Huggins2019-4,Chen2020_11,Dborin2022-7,Miao2021_08,Dilip2022-4,Guala2023-13}.
  \item 
  Neural networks can encounter the vanishing gradient problem when increasing the number of layers \cite{Bengio1993-3,Hochreiter1998-06,Fukumizu2000-13,Dauphin2014-2,Shalev2017-70}.
  This problem may be eliminated in TTN and MERA classifiers as their only nonlinear elements are in the input and output layers.
  In fact, recent results on isometric tensor networks \cite{Barthel2023_03,Miao2024-109,Liu2022-129} suggest that vanishing gradients should not occur for TTN, MPS, and MERA classifiers when the features to be learned have certain locality properties.
  \item
  The number of parameters and computation costs for a full-rank TTN classifier with bond dimension $m$ and tree branching ratio $b$ scale as 
  $\mc{O}(m^{b+1})$ [Eq.~\eqref{eq:costFull}]. The CP rank constraints substantially reduce the number of parameters and costs to $\mc{O}(b m r)$ [Eq.~\eqref{eq:costLowRank}] for the low-rank TTN classifier, where $r$ is an upper bound on the CP rank of the tensors composing the TTN.
  The rank constraints mitigate overfitting and improve the generalization properties. Furthermore, they allow us to employ trees with large branching ratios. This implies that the average graph distance of inputs reduces, which we find to substantially improve the representation power of the classifier.
  Finally, the CP decomposed form of the tensors allows us to employ the tensor dropout method \cite{Kolbeinsson2021-15}, which can further improve the generalization properties.
\end{enumerate}

Section~\ref{sec:TTN_classifier} explains the full and low-rank TTN methods for image classification. Benchmark computations for the MNIST and Fashion-MNIST datasets in Secs.~\ref{sec:MNIST} and \ref{sec:FashMNIST} demonstrate advantages of the low-rank TTN. With low computation costs it outperforms the full TTN, other MPS and PEPS-based classifiers, as well as classification methods like XGBoost \cite{Chen2016_08,Xiao2017_08} and AlexNet \cite{Krizhevsky2017-60,Xiao2017_08}, reaching a Fashion-MNIST test set accuracy of 90.3\% for branching ratio $b=4$. CP ranks $r\gtrsim m$ are found to be sufficient.
There is ample room for further improvements. We conclude and describe some interesting research directions in Sec.~\ref{sec:discuss}.

\section{Image classification with low-rank TTN}\label{sec:TTN_classifier}
\begin{figure*}[t]
    \includegraphics[width=\textwidth]{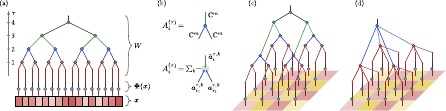}
    \caption{\label{fig:TTN_classifier} A TTN classifier takes an image $\vx\in\RR^N$ as input, transforms it to a feature vector $\vec{\Phi}(\vx)$ in a much higher-dimensional space $(\CC^{2})^{\otimes N}$, and then applies the weight tensor $W:(\CC^{2})^{\otimes N}\to\CC^L$, realized by a TTN, to obtain the decision function \eqref{eq:decisionFct}. (a) A TTN classifier with tree branching ratio $b=2$ for 1D images. (b) Tensors $A^{(\tau)}_i$ of intermediate layers are linear maps from $(\CC^m)^{\otimes b}$ to $\CC^m$. In low-rank TTN classifiers, we reduce computation costs and improve generalization properties by imposing constraints \eqref{eq:CPD} on the CP ranks of the tensors. (c) 2D image TTN classifier with branching ratio $b=2$. (d) 2D image TTN classifier with $b=4$. See, for example, Refs.~\cite{Orus2014-349,Stoudenmire2016-29} for details on the employed graphical representation for tensor networks.}
\end{figure*}
We implement a tree tensor network (TTN) classifier with low-rank tensors for image classification.
Let each image $\vx=(x_1,\dotsc,x_N)^\intercal$ consist of $N$ grayscale pixels $x_j\in[0,1]$. Our task is to assign every image to one of $L$ classes based on corresponding training data
\begin{equation}\label{eq:trainingSet}
	\mc{D}=\{(\vx_s,\ell_s)\}\quad\text{with}\quad\ell_s\in\{1,\dotsc,L\}
\end{equation}
being the class label for image $\vx_s$.

\subsection{Model and loss function}
To facilitate the classification, the images $\vx$ are first mapped into a high-dimensional vector space.
Specifically, we map each pixel value $x_j$ onto a two-dimensional \emph{pixel feature vector} through the local feature map \cite{Stoudenmire2016-29}
\begin{equation}\label{eq:featureMap}
    \vphi(x_j):=\Pmatrix{ \cos(\pi x_j/2)\\ \sin(\pi x_j/2) }
\end{equation}
such that an image $\vx$ is mapped to the tensor product
\begin{equation}\label{eq:featureMapFull}
	\vec{\Phi}(\vx)=\vphi(x_1)\otimes\vphi(x_2)\otimes\dotsb\otimes\vphi(x_N)\in(\CC^{2})^{\otimes N}.
\end{equation}
We then apply a linear map $W:\CC^{2^N}\to\CC^L$, the \emph{weight tensor}, to the image feature vector $\vec{\Phi}(\vx)$ to get the \emph{decision function}
\begin{equation}\label{eq:decisionFct}
	\vf(\vx):=W\cdot \vec{\Phi}(\vx)\in\CC^L.
\end{equation}

The $L$-dimensional output $\vf(\vx)$ is interpreted as an unnormalized quantum state, which is used to compute the probability of image $\vx$ belonging to class $\ell$ via Born's rule
\begin{equation}
	p(\ell|\vx) = |f_\ell(\vx)|^2/\|\vf(\vx)\|^2,
\end{equation}
where $f_\ell$ denotes the $\ell^\text{th}$ component of $\vf$.
The class label of image $\vx$ is then predicted by
\begin{equation}\label{eq:prediction}
	\hat{\ell}(\vx) = \operatorname{argmax}_{1\leq \ell\leq L} \, p(\ell|\vx).
\end{equation}
For the optimization of $W$, we follow the standard approach in supervised learning, minimizing the negative log-likelihood \cite{Mohri2018,Murphy2013} for the training set \eqref{eq:trainingSet},
\begin{equation}\label{eq:NLL}
	   \mc{L} = - \frac{1}{|\mc{D}|} \sum_{(\vx_s, \ell_s) \in \mc{D}} \ln \,p(\ell_s|\vx_s).
\end{equation}

\subsection{TTN weight tensor and contractions}
Similar to approaches in Refs.~\cite{Liu2019-21,Cheng2019-99,Stoudenmire2018-3,Grant2018-4,Huggins2019-4}, we implement the weight tensor $W$ as a TTN. The underlying tree graph is organized in layers and characterized by a \emph{branching ratio} $b$ and \emph{bond dimension} $m$. See Fig.~\ref{fig:TTN_classifier}. The lowest layer consists of tensors $A^{(1)}_i\in\CC^{m\times 2^b}$ with $i=1,\dotsc,N/b$, each mapping the pixel feature vectors \eqref{eq:featureMap} of $b$ pixels $i_1,\dotsc,i_b$ into a \emph{level-1 feature vector} of dimension $m$,
\begin{subequations}
\begin{equation}
	\vvphi^{(1)}_i=A^{(1)}_i\cdot\big(\vphi(x_{i_1})\otimes\dotsb\otimes\vphi(x_{i_b})\big)\in\CC^m,
\end{equation}
where we interpreted tensor $A^{(1)}_i$ as a linear map from $(\CC^2)^{\otimes b}$ to $\CC^{m}$, i.e., the $\mu$ component of $\vvphi^{(1)}_i$ is
\begin{equation}
	\varphi^{(1)}_{i,\mu}=\sum_{s_1,\dotsc,s_b=1}^2 [A^{(1)}_i]_{\mu,s_1,\dotsc,s_b}\,\,\phi_{s_1}(x_{i_1})\dotsb\phi_{s_b}(x_{i_b}).
\end{equation}
\end{subequations}
Similarly, tensors $A^{(\tau)}_i\in\CC^{m\times m^b}$ of layer $\tau>1$ each map $b$ level-$(\tau-1)$ feature vectors into a level-$\tau$ feature vector of dimension $m$,
\begin{subequations}\label{eq:contractMiddle}
\begin{equation}
	\vvphi^{(\tau)}_i=A^{(\tau)}_i\cdot\big(\vvphi^{(\tau-1)}_{i_1}\otimes\dotsb\otimes\vvphi^{(\tau-1)}_{i_b}\big)\in\CC^m,
\end{equation}
where we interpreted tensor $A^{(\tau)}_i$ as a linear map from $(\CC^m)^{\otimes b}$ to $\CC^{m}$, i.e., the $\mu$ component of $\vvphi^{(\tau)}_i$ is
\begin{equation}
	\varphi^{(\tau)}_{i,\mu}=\sum_{\mu_1,\dotsc,\mu_b=1}^m [A^{(\tau)}_i]_{\mu,\mu_1,\dotsc,\mu_b}\,\,\varphi^{(\tau-1)}_{i_1,\mu_1}\dotsb\varphi^{(\tau-1)}_{i_b,\mu_b}.
\end{equation}
\end{subequations}
The procedure ends in layer $T=\log_b N$ with the top tensor $A^{(T)}\in\CC^{L\times m^b}$ mapping the remaining $b$ level-$(T-1)$ feature vectors into the $L$-dimensional vector \eqref{eq:decisionFct},
\begin{equation}
	\vf(\vx)=A^{(T)}\cdot\big(\vvphi^{(T-1)}_{1}\otimes\dotsb\otimes\vvphi^{(T-1)}_{b}\big)\in\CC^L.
\end{equation}
So, the weight tensor $W$ is chosen as a network of smaller tensors $A^{(1)}_i,A^{(2)}_i,\dotsc,A^{(T)}$, each assigned to a vertex of a tree graph and 
all except for the input indices of the tensors $A^{(1)}_i$ and the output index of $A^{(T)}$ being contracted. Each edge of the graph is associated with a common index $\mu$ of the two tensors of the corresponding vertices, which we sum over in the tensor contractions.
The minimization of the loss function \eqref{eq:NLL} with respect to the elements of the tensors $\{A^{(\tau)}_i\}$ can be accomplished through gradient-based optimization methods.

Figure~\ref{fig:TTN_classifier} shows three examples. For a one-dimensional image with $N=2^T$ pixels, each layer of a TTN classifier with branching ratio $b=2$ halves the number of feature vectors (Figure~\ref{fig:TTN_classifier}a).
For a two-dimensional image with $N=N_x\times N_y=2^T$ pixels, a TTN classifier with branching ratio $b=2$, can alternatingly coarse-grain in the $x$ and $y$ direction, each layer halving the number of feature vectors as illustrated in Fig.~\ref{fig:TTN_classifier}c. Similarly, for $N=N_x\times N_y=2^T\times 2^T$, each layer of a TTN classifier with branching ratio $b=4$, can simultaneously coarse-grain both directions, mapping $2\times 2$ patches into one feature vector as shown in Fig.~\ref{fig:TTN_classifier}d. The approach is very flexible and one can, for example, work with networks that contain tensors of varying order.

When we do not impose any constraints on the tensors $A^{(\tau)}_i$, we refer to the method as a full TTN classifier. The number of parameters and the cost for the loss function evaluation then scale as
\begin{equation}\label{eq:costFull}
	\mc{O}(N m^{b+1}) \quad \text{and}\quad
	\mc{O}(|\mc{D}|N m^{b+1}),
\end{equation}
respectively.
So, the larger the bond dimension $m$, the higher the expressiveness (representation power) of the TTN classifier and the higher the computation costs. We will also see how large $m$ can result in overfitting. Generally, larger branching ratios $b$ have the advantage of reducing the graph distance of pixels in the tensor network. When keeping $m$ constant, larger $b$ allows for a more efficient extraction of non-local class features.

\subsection{TTN with low-rank tensors and tensor dropout}
The main contribution of this work is to improve tensor-network machine learning by introducing more flexibility concerning the number of parameters and substantially reducing computation costs. The idea, which we will demonstrate for the TTN classifiers, is to impose a constraint on the \emph{canonical polyadic (CP) rank} of all tensors in the tensor network. Similarly, we have recently introduced and tested CP rank constraints for tensor network state simulations of quantum systems \cite{Chen2022_05}.

The canonical polyadic decomposition \cite{Hitchcock1927-6,Carroll1970-35,Harshman1970-16,Kolda2009-51} is a generalization of the singular value decomposition to higher-order tensors, which expresses a tensor $A^{(\tau)}_i$ as the sum of direct products of vectors. For example, every tensor $A^{(\tau)}_i\in\CC^{m\times m^b}$ from one of the intermediate layers $1<\tau<T$ can be written in the form
\begin{equation}\label{eq:CPD}
	 A^{(\tau)}_i = \sum_{k=1}^r \tilde{\va}^{\tau,k}_i\otimes \va^{\tau,k}_{i_1}\otimes \dotsb \otimes \va^{\tau,k}_{i_b},
\end{equation}
with order-1 tensors (vectors) $\tilde{\va}^{\tau,k}_i,\va^{\tau,k}_{i_1},\dotsc,\va^{\tau,k}_{i_b} \in \CC^{m}$. The minimal $r$ satisfying Eq.~\eqref{eq:CPD} is called the \emph{canonical polyadic rank} or \emph{candecomp-parafac rank} of $A^{(\tau)}_i$.

In a \emph{low-rank TTN classifier}, we choose every tensor to be of the form \eqref{eq:CPD} with a fixed $r$. The same form is used for the input-layer tensors $A^{(1)}_i$ with $\va^{1,k}_{i_1},\dotsc,\va^{1,k}_{i_b} \in \CC^{2}$ and for the top tensor $A^{(T)}_i$ with $\tilde{\va}^{T,k} \in \CC^{L}$.
The rank constraint allows us to execute contractions much more efficiently. In particular, the contraction \eqref{eq:contractMiddle} can be implemented by first computing 
\begin{subequations}
\begin{equation}
	\psi^{\tau,k}_{i_n}=\sum_{\mu=1}^m \va^{\tau,k}_{i_n}\varphi^{(\tau-1)}_{i_n,\mu}
\end{equation}
for $n=1,\dotsc,b$ and $k=1,\dots,r$ and, then, evaluating
\begin{equation}
	\vvphi^{(\tau)}_i=\sum_{k=1}^r \tilde{\va}^{\tau,k}_i\,\psi^{\tau,k}_{i_1}\dotsb\psi^{\tau,k}_{i_b}.
\end{equation}
\end{subequations}
So, the number of parameters and the cost for the loss function evaluation reduce from Eq.~\eqref{eq:costFull} for the full-rank TTN to
\begin{equation}\label{eq:costLowRank}
	\mc{O}(N b m r ) \quad \text{and}\quad
	\mc{O}(|\mc{D}|N b m r)
\end{equation}
for the low-rank TTN.

Lastly, the low-rank TTN classifier can be further regularized, improving its generalization properties through \emph{tensor dropout} \cite{Kolbeinsson2021-15}. The basic idea is to randomly suppress some of the terms in the CP form \eqref{eq:CPD} of the tensors during supervised learning. Details of this procedure are described in Sec.~\ref{sec:FashMNIST}.

\section{MNIST benchmark computations}\label{sec:MNIST}
\begin{figure}[t]
    \centering
    \includegraphics[width=0.65\columnwidth]{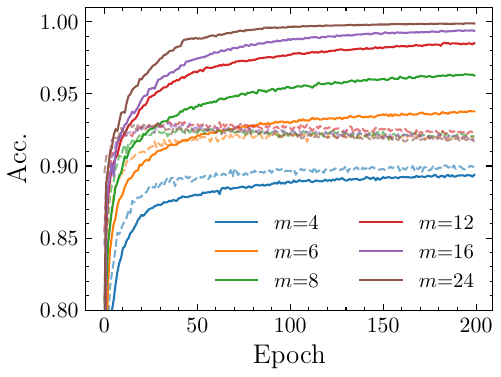}
    \caption{MNIST training histories of full TTN classifiers with branching ratio $b=4$ and a few different bond dimensions $m$.
    Solid and dashed lines represent the classification accuracy \eqref{eq:accuracy} for the training set and validation set, respectively.}
    \label{fig:mnist_full}
\end{figure}
\begin{figure}[t]
    \includegraphics[width=\columnwidth]{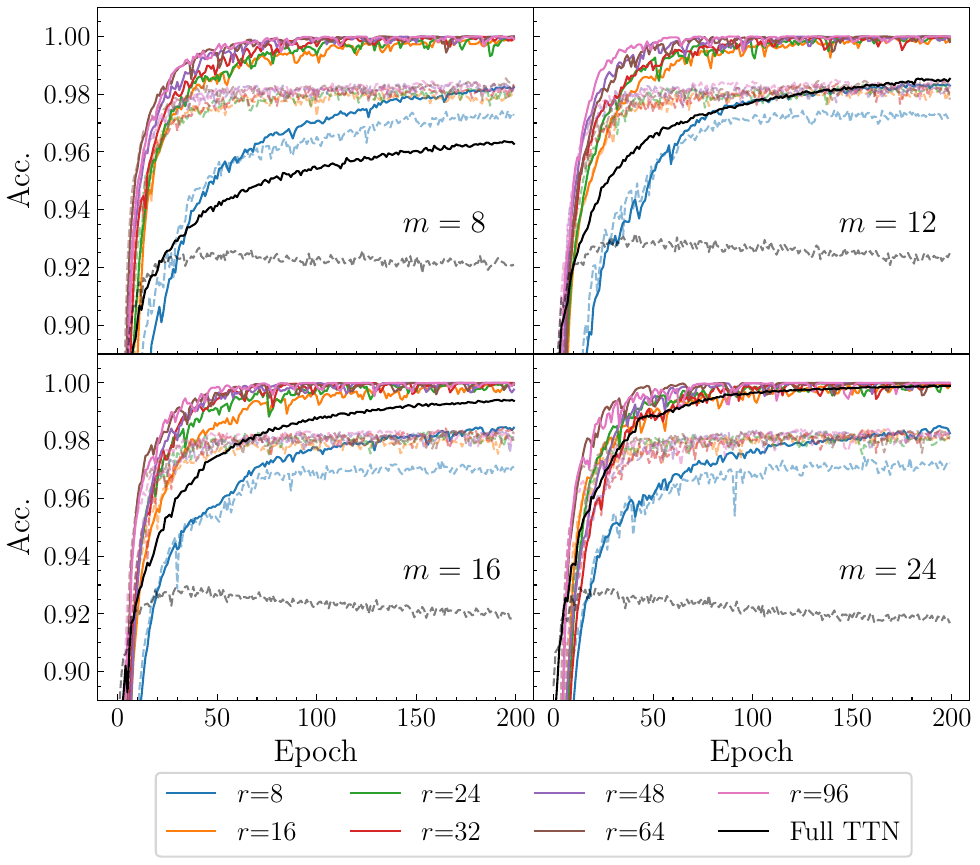}
    \caption{MNIST training histories of low-rank TTN classifiers with branching ratio $b=4$, varying bond dimensions $m$, and CP ranks $r$.
    Solid and dashed lines represent the classification accuracy \eqref{eq:accuracy} for the training set and validation set, respectively.
    For comparison, the black lines show the full TTN results of Fig.~\ref{fig:mnist_full}.}
    \label{fig:mnist_0.0}
\end{figure}
We first compare full and low-rank TTN classifiers with branching ratio $b=4$ for the MNIST dataset of handwritten digits \cite{Lecun1998-86}. It consists of $28\times 28$ grayscale images falling into $L = 10$ classes. The database is split into \numprint{55000} training images, \numprint{5000} validation images, and \numprint{10000} test images. We downsize each image to $16\times 16$ pixels so that it can be easily fed into a TTN classifier with branching ratio  $b=2$ and $T=8$ layers or $b=4$ and $T=4$. For MNIST, we will employ the latter option. Instead of rescaling the images, which leads to some information loss, one could use $b=4$ with $T=5$ layers and pad the images with white pixels to a $32\times 32$ format or use a TTN with a non-uniform branching ratio.

We initialize tensor elements of full TTN tensors $A^{\tau}_i$ and the elements of the low-rank TTN vectors $\tilde{\va}^{\tau,k}_i,\va^{\tau,k}_{i_1},\dotsc,\va^{\tau,k}_{i_b}$ in Eq.~\eqref{eq:CPD} by sampling real and imaginary parts from a normal distribution with mean zero and a standard deviation of $0.4$. The training is performed using the Adam optimizer \cite{Kingma2014_12,Abadi2015} with an initial learning rate of $\alpha=0.001$ and decay factors $\beta_1=0.9$ and $\beta_2=0.999$ for the first-moment and second-moment estimates. The parameter $\hat{\epsilon}$, which prevents divisions by zero, is set to $\hat{\epsilon}=10^{-7}$. See also Refs.~\cite{Loshchilov2017_11,Reddi2019_04}.

Figures~\ref{fig:mnist_full} and \ref{fig:mnist_0.0} show histories of classification accuracies
for full TTN and low-rank TTN, respectively. The accuracy is defined as
\begin{equation}\label{eq:accuracy}
	\text{Acc.}\stackrel{\eqref{eq:prediction}}{=}\frac{1}{|\mc{D}'|} \sum_{(\vx_s, \ell_s)\in\mc{D}'} \delta_{\ell_s,\hat{\ell}(\vx_s)}.
\end{equation}
Here, $\delta_{\ell,\ell'}$ is the Kronecker delta ($\delta_{\ell,\ell'}=1$ for $\ell=\ell'$ and zero otherwise), and $D'$ refers either to the training, validation, or test dataset.
The accuracies are shown for various bond dimensions $m$. For the low-rank TTN, we also vary the CP rank $r$. Solid and dashed lines represent the training and validation set accuracy, respectively.

For the full TTN, the expressiveness of the classifier increases with increasing bond dimension $m$, as reflected by the improving training set accuracy. However, the model does not generalize well on the validation set as classifiers with $m > 4$ all give approximately 92-93\% validation set accuracy, well below the best training set accuracies. This is a typical feature of overfitting. In contrast, the low-rank TTN provide an implicit regularization, and the low-rank TTN classifiers are able to achieve
98.3\% test accuracy. Importantly, the CP rank $r$ does not play a pivotal role in the performance as long as $r > m$. Similar to observations for low-rank tensor network state simulations of quantum systems in Ref.~\cite{Chen2022_05}, $r\approx m$ seems sufficient for the low-rank TTN classifiers. In addition to the improved generalization properties, according to Eqs.~\eqref{eq:costFull} and \eqref{eq:costLowRank}, this implies substantial computation cost reductions.

\section{Fashion-MNIST benchmark computations}\label{sec:FashMNIST}
As a second test, we compare full and low-rank TTN classifiers with branching ratios $b=2$ and $b=4$ for the more challenging Fashion-MNIST dataset \cite{Xiao2017_08}. It comprises \numprint{70000} grayscale $28\times 28$ images for $L = 10$ classes of clothing. We split the dataset in the same way as for MNIST and apply the same methods for image rescaling and TTN initialization.

\begin{figure}[t]
    \centering
    \includegraphics[width=0.63\columnwidth]{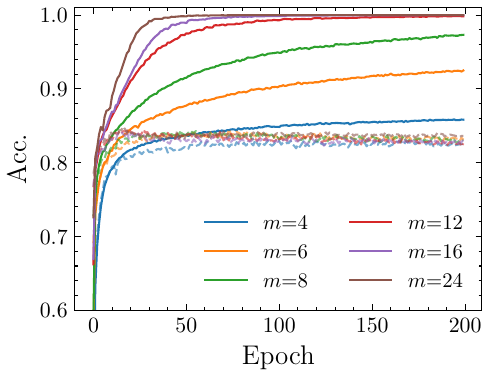}
    \caption{Fashion-MNIST training histories of full TTN classifiers with branching ratio $b=4$ and a few different bond dimensions $m$.
    Solid and dashed lines represent the classification accuracy \eqref{eq:accuracy} for the training set and validation set, respectively.}
    \label{fig:fmnist_full}
\end{figure}

\begin{figure}[t]
    \includegraphics[width=0.99\columnwidth]{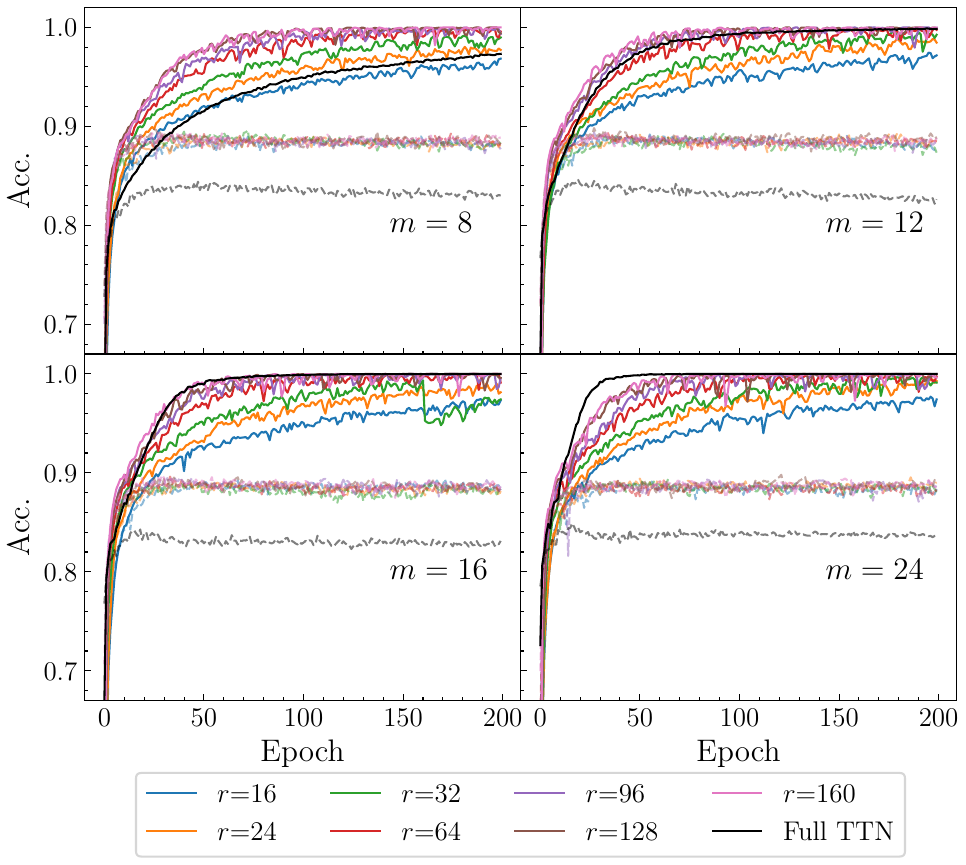}
    \caption{Fashion-MNIST training histories of low-rank TTN classifiers with branching ratio $b=4$, varying bond dimensions $m$, and CP ranks $r$.
    Solid and dashed lines represent the classification accuracy \eqref{eq:accuracy} for the training set and validation set, respectively.
    For comparison, the black lines show the full TTN results of Fig.~\ref{fig:fmnist_full}.}
    \label{fig:fmnist_0.0}
\end{figure}

Training histories for full and low-rank TTN classifiers with branching ratio $b=4$ and $T=4$ layers are presented in Figs.~\ref{fig:fmnist_full} and \ref{fig:fmnist_0.0}. We find that the overfitting issue of the full TTN classifier is more prominent for Fashion-MNIST. As shown in Fig.~\ref{fig:fmnist_full}, although networks with $m \geq 12$ can reach nearly 100\% training set accuracy, their accuracies on the validation set reach only about 84\%.
Similar to the MNIST case, introducing the CP rank constraint on the TTN tensors substantially improves the generalization properties of the network. The low-rank TTN classifier achieves 88.5\% classification accuracy on the test dataset. 

Even for low-rank TTN, there is apparently still a considerable degree of overfitting. For almost all considered $m$ and $r$, the training set accuracy converges to almost 100\%, but the validation set accuracy converges (much more quickly) to a value close to the aforementioned test set accuracy of 88.5\%.

To further improve the generalization properties, we leverage the CP form \eqref{eq:CPD} of the rank-constrained tensors and
employ the \emph{tensor dropout} method proposed in Ref.~\cite{Kolbeinsson2021-15} based on similar approaches used in neural-network training \cite{Srivastava2014-15}:
In the training phase, each component in the sum of Eq.~\eqref{eq:CPD} is associated with a Bernoulli random variable $\lambda^{\tau,k}_i$ taking value 1 with probability $1-p$ and 0 with probability $p$. Only the terms with $\lambda^{\tau,k}_i=1$ are retained, and the new tensors after the dropout operation are chosen as
\begin{equation}\label{eq:droputA}
	 \tilde{A}^{(\tau)}_i = \frac{1}{1-p}\, \sum_{k=1}^r \lambda^{\tau,k}_i\, \tilde{\va}^{\tau,k}_i\otimes \va^{\tau,k}_{i_1}\otimes \dotsb \otimes \va^{\tau,k}_{i_b}.
\end{equation}
We call $p$ the \emph{dropout rate}. For each training iteration, the $\lambda^{\tau,k}_i$ are sampled from the Bernoulli distribution, the outputs $\vf(\vx_s)$ of the network are computed using the modified tensors \eqref{eq:droputA}, and the tensor parameters are updated accordingly using backpropagation algorithms. 
When the training is completed or its classification accuracy is tested, the tensor dropout operation is removed ($p=0$). In our implementation, we apply tensor dropout on all tensors except for the top tensor $A^{(T)}$.
\begin{figure}[t]
    \includegraphics[width=\columnwidth]{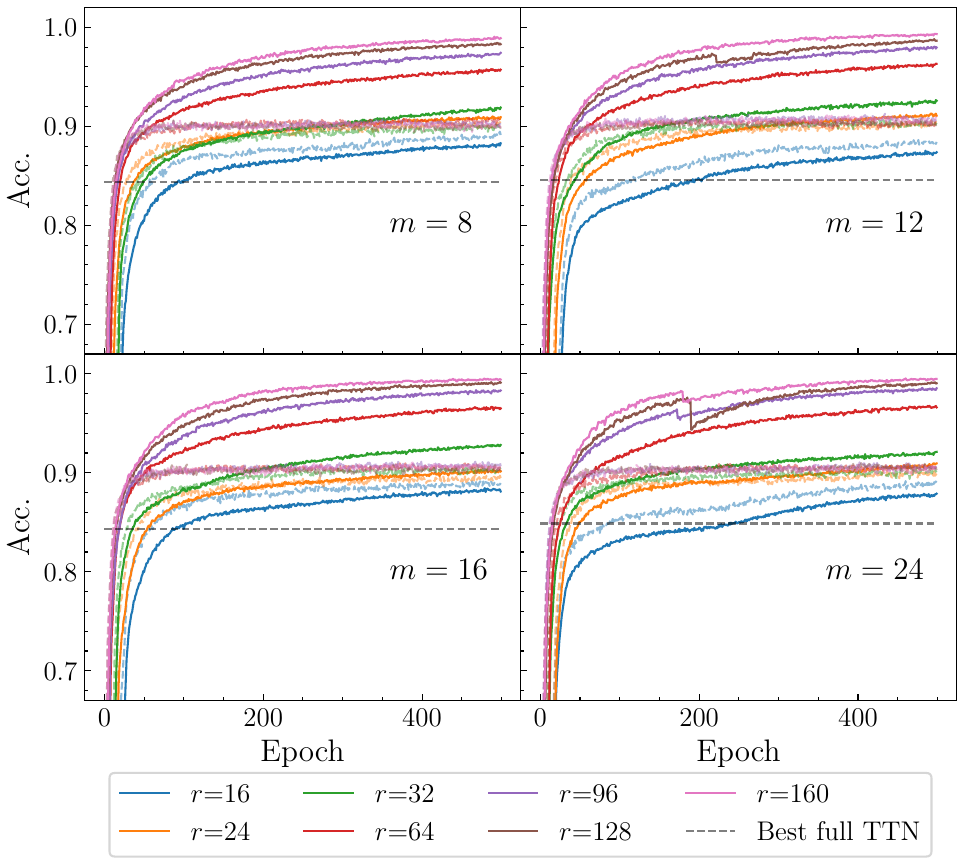}
    \caption{Fashion-MNIST training histories of low-rank TTN classifiers with tensor dropout rate $p=0.2$ and branching ratio $b=4$.
    Solid and dashed lines are the classification accuracy \eqref{eq:accuracy} for the training set and validation set, respectively. For comparison, dashed horizontal lines indicate the best validation accuracy of the full TTN from Fig.~\ref{fig:fmnist_full}.}
    \label{fig:fmnist_0.2}
\end{figure}

Figure~\ref{fig:fmnist_0.2} shows training histories for low-rank TTN classifiers with branching ratio $b=4$ and the tensor dropout regularization with rate $p=0.2$. Compared to the results without tensor dropout, training set accuracies now converge more slowly and display smaller fluctuations. Moreover, the classification accuracies for the validation set are improved by about 1-2\%.

\begin{figure}[t]
    \includegraphics[width=\columnwidth]{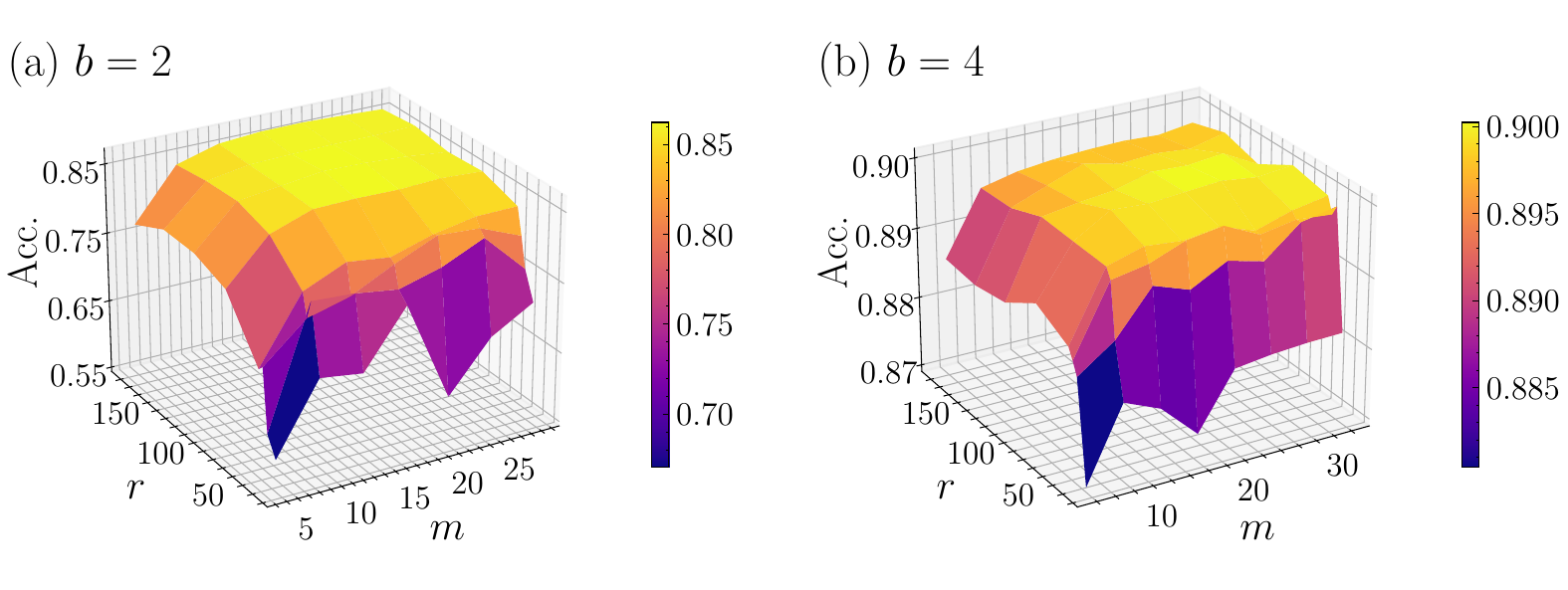}
    \caption{Dependence of the Fashion-MNIST test set accuracy on the bond dimension $m$ and CP rank $r$ for low-rank TTN classifiers with branching ratio $b=2$ (left) and $b=4$ (right). The tensor dropout rate was set to $p=0.2$.
    To mitigate fluctuations of the test accuracy, at each point, we have averaged the results of five runs with different initial conditions.
    In both cases, the accuracy rapidly reaches a plateau, but the $b=4$ classifier clearly outperforms the one with $b=2$. Note the different accuracy scales and color scales in the two plots.}
    \label{fig:acc_r_m}
\end{figure}

Figure~\ref{fig:acc_r_m}b shows the dependence of the test set accuracy on the bond dimension $m$ and CP rank $r$ for the low-rank TTN with $b=4$ and dropout rate $p=0.2$. The classification accuracy increases rapidly when $m$ or $r$ is small and enters an extended plateau for $m \gtrsim 12$ and $r\gtrsim 32$. The best test set accuracy of 90.3\% observed in these simulations occurs for $m=16$, $r=64$, and $p=0.3$.

To study the effect of the branching ratio $b$ on the classification accuracy of the low-rank TTN, we perform similar experiments with $b=2$. Compared to the $b=4$ networks, the number of layers doubles to $T=8$. During the training of TTN with large $m$ and $r$, we encountered some diverging vector components for the tensors \eqref{eq:CPD}. This can be explained by the fact that, for fixed network structure, $m$, and $r$, the set of low-rank TTN classifiers is generally not closed \cite{Bini1979-8,DeSilva2008-30,Barthel2022-112}. To avoid convergence to non-included boundary points of the set, which is associated with diverging vector elements,
we add the penalty term $\gamma \sum_{\tau,i} \|A^{(\tau)}_i\|^2$ to the loss function \eqref{eq:NLL} as suggested in Ref.~\cite{Barthel2022-112}. Here, we choose $\gamma=0.01$. The results are shown in Fig.~\ref{fig:acc_r_m}a. While the test set accuracy as a function of $m$ and $r$ displays a plateau similar to the case with branching ratio $b=4$, the test accuracy for $b=2$ is generally lower by around 4\%.
\begin{table}[b]
\caption{Test set accuracies of different image classifiers for the Fashion-MNIST dataset.}\label{tab:fashion_mnist_test_acc}
\begin{tabular}{lc}
    \hline
    Model  &\multicolumn{1}{c}{\ Test Accuracy} \\
    \hline
    MPS\cite{Efthymiou2019_06}	& 88.0\% \\
    PEPS \cite{Cheng2021-103}	& 88.3\% \\
    \textbf{Low-rank TTN}	& 88.5\% \\
    MPS+TTN\cite{Stoudenmire2018-3} & 89.0\% \\
    XGBoost\cite{Xiao2017_08}	& 89.8\% \\ 
    AlexNet\cite{Xiao2017_08}	& 89.9\% \\
    \textbf{Low-rank TTN (with tensor dropout)} & 90.3\% \\
    GoogLeNet\cite{Xiao2017_08}	& 93.7\% \\
   \hline
\end{tabular}
\end{table}

This can be attributed to the larger average graph distance of pixels in the TTN with smaller branching ratio $b$. As an example, features that depend on both the left and right halves of the image can only be detected by the top tensor, i.e., after seven layer transitions for $b=2$ but already after three layer transitions for $b=4$. The data in Fig.~\ref{fig:acc_r_m} confirms the benefits of large $b$, which can be achieved much more easily with the low-rank TTN classifiers.
\begin{figure}[t]
    \centering
    \includegraphics[width=0.7\columnwidth]{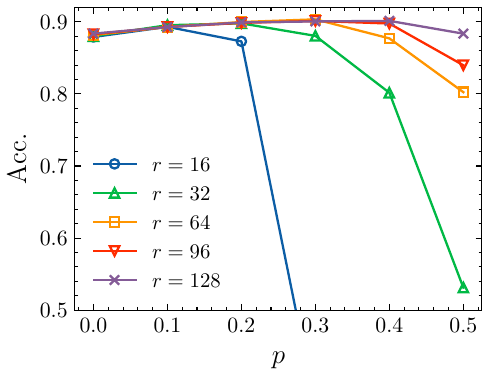}
    \caption{The Fashion-MNIST test set accuracy as a function of tensor dropout rate $p$ for various CP ranks $r$. Here we use branching ratio $b=4$ and bond dimension $m=16$.}
    \label{fig:dropout_rates}
\end{figure}

Figure~\ref{fig:dropout_rates} assesses the dependence of the test set accuracy on the dropout rate $p$. Starting from $p=0$, the curves for fixed $m=16$ and several $r$ all display a slow increase in the accuracy, which is followed by a more rapid decrease at larger $p$. The turning point $p_c$ increases with increasing CP rank $r$, and networks with higher ranks are more robust in the sense that the final accuracy decrease for $p > p_c$ is not as drastic.

Finally, we compare our results with other state-of-the-art classification methods in Table~\ref{tab:fashion_mnist_test_acc}. The low-rank TTN model with tensor dropout outperforms other tensor-network-based approaches. Despite its simple structure and low computation costs, the low-rank TTN with tensor dropout is also competitive with more prominent machine-learning approaches like the convolutional-neural-network-based AlexNet \cite{Krizhevsky2017-60} and the decision-tree based XGBoost \cite{Chen2016_08}.

\section{Discussion}\label{sec:discuss}
The only nonlinear steps in the TTN classifiers are the feature map \eqref{eq:featureMap} and the class label prediction \eqref{eq:prediction}. Also supported by the benchmark simulations, we expect that there is hence no vanishing gradient problem as encountered for neural networks \cite{Bengio1993-3,Hochreiter1998-06,Fukumizu2000-13,Dauphin2014-2,Shalev2017-70}. In fact, it has recently been shown that energy minimization problems for isometric tensor network optimization are free of barren plateaus for systems with finite-range interactions \cite{Barthel2023_03,Miao2024-109,Liu2022-129}. It is conceivable that this result can be extended to the image classification problem. To this purpose, one needs to assess locality properties of the corresponding loss functions such as the negative log-likelihood \eqref{eq:NLL} or the structure of correlations in the optimized classifier, e.g., by studying entanglement properties \cite{Liu2019-21,Martyn2020_07,Liu2021-7,Convy2022-3a,Dymarsky2022-4}.

On the other hand, having fewer non-linear elements may of course also reduce the expressiveness or generalization power. It would hence be interesting to combine the low-rank TTN weight tensor with more elaborate nonlinear feature maps. For example, it would be natural to employ $d$-dimensional feature vectors
\begin{equation*}
    \phi_s(x_j)=\sqrt{\Pmatrix{ d-1 \\ s_j-1}}\left(\cos \frac{\pi x_j}{2}\right)^{d-s_j}\left(\sin \frac{\pi x_j}{2}\right)^{s_{j}-1},
\end{equation*}
with $s=1,\dotsc,d$ as suggested in \cite{Stoudenmire2016-29}. Here, we can increase from $d=2$, corresponding to Eq.~\eqref{eq:featureMap}, to $d\approx m$ without increasing the overall scaling of computation costs [Eqs.~\eqref{eq:costFull} and \eqref{eq:costLowRank}].

Another simple route for improvements would be to employ a separate TTN for every class, i.e., to work with $L$ binary classifiers where the top tensor maps into $\CC^2$ instead of $\CC^L$ as done in Ref.~\cite{Liu2019-21}.

Both for MNIST and Fashion-MNIST, we observed that choosing CP ranks to be on the order of the bond dimension, $r\gtrsim m$, is sufficient to reach high classification accuracies and that the accuracies did not improve substantially for larger $r$. This is consistent with observations in low-rank TTN simulations of quantum many-body systems \cite{Chen2022_05}. It would be valuable to explain this phenomenon, which might be possible by analyzing exactly solvable problems.

A major advantage of the CP rank constraints in the low-rank TTN is that we can efficiently accommodate much larger tree branching ratios $b$, compared to full TTN. This reduces the mutual graph distance of pixels in the tensor network and, hence, allows for a more efficient representation of correlations.
In the presented benchmark simulations we only used $b\leq 4$ because of the small MNIST and Fashion-MNIST image sizes. We observed substantial accuracy gains when going from $b=2$ to $b=4$. It will hence be interesting to investigate the performance of low-rank TTN classifiers with even larger $b$ in more challenging machine learning problems.
In the current study, we found that allowing for complex TTN tensor elements reduced issues with local minima and, generally, leads to more robust convergence properties. Currently, we do not have an obvious explanation and the issue deserves further analysis.

The basic idea of tensor network classifiers is to map the input into a large vector space that is exponentially big in the number of input variables. Tensor networks are then employed to efficiently encode the high-dimensional weight tensor. The high dimensionality helps in separating classes and, using different cost functions, the same logic applies for generative modeling and clustering problems \cite{Han2018-8,Stoudenmire2018-3,Cheng2019-99,Sun2020-101,Bai2022-39,Shi2022-105,Fernandez2022-12,Vieijra2022_02,Lidiak2022_07,Liu2023-107,Shi2023_02}.

\section*{Acknowledgment}
We gratefully acknowledge discussions with participants of the IPAM program \emph{``Tensor methods and emerging applications to the physical and data sciences''} (2021--2023).

\begin{minipage}{\columnwidth}
\begin{IEEEbiography}[{\includegraphics[width=1in,clip,keepaspectratio]{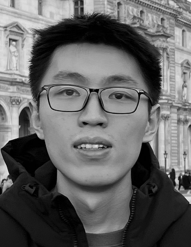}}]{Hao Chen}
currently pursues MSc.\ studies in physics at ETH Zurich, Switzerland. He received a bachelor's degree of science in applied physics from the University of Science and Technology of China. His research interests are in quantum many-body physics, computational physics, and machine learning.
\end{IEEEbiography}
\begin{IEEEbiography}[{\includegraphics[width=1in,clip,keepaspectratio]{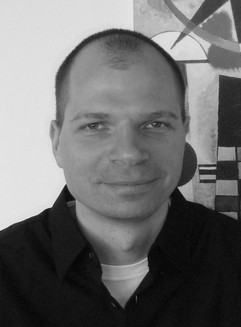}}]{Thomas Barthel}
is a professor of physics at Duke University and member of the Duke Quantum Center. He received a PhD degree in physics from RWTH Aachen University and held postdoctoral positions at University of Potsdam, FU Berlin, LMU Munich, and Universit\'{e} Paris-Saclay. His research interests cover theoretical and computational quantum many-body physics, tensor network state methods, quantum information theory, and machine learning.
\end{IEEEbiography}
\end{minipage}


\begin{thebibliography}{10}

\bibitem{Orus2014-349}
R. Or\'{u}s, {\em A practical introduction to tensor networks: Matrix product
  states and projected entangled pair states},
  \href{https://doi.org/10.1016/j.aop.2014.06.013} {Ann. Phys. {\bf 349},  117
   (2014)}.

\bibitem{Baxter1968-9}
R.~J. Baxter, {\em Dimers on a rectangular lattice},
  \href{https://doi.org/10.1063/1.1664623} {J. Math. Phys. {\bf 9},  650
  (1968)}.

\bibitem{Fannes1992-144}
M. Fannes, B. Nachtergaele, and R.~F. Werner, {\em Finitely correlated states
  on quantum spin chains}, \href{https://doi.org/10.1007/BF02099178} {Commun.
  Math. Phys. {\bf 144},  443  (1992)}.

\bibitem{White1992-11}
S.~R. White, {\em Density matrix formulation for quantum renormalization
  groups}, \href{https://doi.org/10.1103/PhysRevLett.69.2863} {Phys. Rev. Lett.
  {\bf 69},  2863  (1992)}.

\bibitem{Rommer1997}
S. Rommer and S. \"Ostlund, {\em A class of ansatz wave functions for 1D spin
  systems and their relation to DMRG},
  \href{https://doi.org/10.1103/PhysRevB.55.2164} {Phys. Rev. B {\bf 55},  2164
   (1997)}.

\bibitem{PerezGarcia2007-7}
D. Perez-Garcia, F. Verstraete, M.~M. Wolf, and J.~I. Cirac, {\em Matrix
  product state representations},
  \href{https://doi.org/10.48550/arXiv.quant-ph/0608197} {Quantum Info. Comput.
  {\bf 7},  401  (2007)}.

\bibitem{Schollwoeck2011-326}
U. Schollw\"{o}ck, {\em The density-matrix renormalization group in the age of
  matrix product states}, \href{https://doi.org/10.1016/j.aop.2010.09.012}
  {Ann. Phys. {\bf 326},  96  (2011)}.

\bibitem{Oseledets2011-33}
I.~V. Oseledets, {\em Tensor-train decomposition},
  \href{https://doi.org/10.1137/090752286} {SIAM J. Sci. Comput. {\bf 33},
  2295  (2011)}.

\bibitem{Fannes1992-66}
M. Fannes, B. Nachtergaele, and R.~F. Werner, {\em Ground states of {VBS}
  models on cayley trees}, \href{https://doi.org/10.1007/bf01055710} {J. Stat.
  Phys. {\bf 66},  939  (1992)}.

\bibitem{Otsuka1996-53}
H. Otsuka, {\em Density-matrix renormalization-group study of the spin-$1/2$
  $\mathrm{XXZ}$ antiferromagnet on the Bethe lattice},
  \href{https://doi.org/10.1103/PhysRevB.53.14004} {Phys. Rev. B {\bf 53},
  14004  (1996)}.

\bibitem{Shi2006-74}
Y.-Y. Shi, L.-M. Duan, and G. Vidal, {\em Classical simulation of quantum
  many-body systems with a tree tensor network},
  \href{https://doi.org/10.1103/PhysRevA.74.022320} {Phys. Rev. A {\bf 74},
  022320  (2006)}.

\bibitem{Hackbusch2009-15}
W. Hackbusch and S. K\"{u}hn, {\em A new scheme for the tensor representation},
  \href{https://doi.org/10.1007/s00041-009-9094-9} {J. Fourier Anal. Appl. {\bf
  15},  706  (2009)}.

\bibitem{Murg2010-82}
V. Murg, F. Verstraete, O. Legeza, and R.~M. Noack, {\em Simulating strongly
  correlated quantum systems with tree tensor networks},
  \href{https://doi.org/10.1103/PhysRevB.82.205105} {Phys. Rev. B {\bf 82},
  205105  (2010)}.

\bibitem{Nakatani2013-138}
N. Nakatani and G.~K.-L. Chan, {\em Efficient tree tensor network states (TTNS)
  for quantum chemistry: Generalizations of the density matrix renormalization
  group algorithm}, \href{https://doi.org/10.1063/1.4798639} {J. Chem. Phys.
  {\bf 138},  134113  (2013)}.

\bibitem{Vidal-2005-12}
G. Vidal, {\em Entanglement renormalization},
  \href{https://doi.org/10.1103/PhysRevLett.99.220405} {Phys. Rev. Lett. {\bf
  99},  220405  (2007)}.

\bibitem{Vidal2006}
G. Vidal, {\em Class of quantum many-body states that can be efficiently
  simulated}, \href{https://doi.org/10.1103/PhysRevLett.101.110501} {Phys. Rev.
  Lett. {\bf 101},  110501  (2008)}.

\bibitem{Niggemann1997-104}
H. Niggemann, A. Kl\"umper, and J. Zittartz, {\em Quantum phase transition in
  spin-3/2 systems on the hexagonal lattice - optimum ground state approach},
  \href{https://doi.org/10.1007/s002570050425} {Z. Phys. B {\bf 104},  103
  (1997)}.

\bibitem{Nishino2000-575}
T. Nishino, K. Okunishi, Y. Hieida, N. Maeshima, and Y. Akutsu, {\em
  Self-consistent tensor product variational approximation for 3D classical
  models}, \href{https://doi.org/10.1016/S0550-3213(00)00133-4} {Nucl. Phys. B
  {\bf 575},  504  (2000)}.

\bibitem{Verstraete2004-7}
F. Verstraete and J.~I. Cirac, {\em Renormalization algorithms for quantum-many
  body systems in two and higher dimensions},
  \href{http://arxiv.org/abs/cond-mat/0407066} {arXiv:cond-mat/0407066
  (2004)}.

\bibitem{Verstraete2006-96}
F. Verstraete, M.~M. Wolf, D. Perez-Garcia, and J.~I. Cirac, {\em Criticality,
  the area law, and the computational power of projected entangled pair
  states}, \href{https://doi.org/10.1103/PhysRevLett.96.220601} {Phys. Rev.
  Lett. {\bf 96},  220601  (2006)}.

\bibitem{Cohen2016-29}
N. Cohen, O. Sharir, and A. Shashua, {\em On the expressive power of deep
  learning: A tensor analysis},
  \href{https://doi.org/10.48550/arXiv.1509.05009} {Ann. Conf. Learn. Theory
  {\bf 29},  698  (2016)}.

\bibitem{Stoudenmire2016-29}
E. Stoudenmire and D.~J. Schwab, {\em Supervised learning with tensor
  networks}, \href{https://doi.org/10.48550/arXiv.1605.05775} {Adv. Neur. Inf.
  Proc. Sys. {\bf 29},  4799  (2016)}.

\bibitem{Novikov2016_05}
A. Novikov, M. Trofimov, and I. Oseledets, {\em Exponential machines},
  \href{https://doi.org/10.48550/arXiv.1605.03795} {International Conference on
  Learning Representations, Workshop Track  (2017)}.

\bibitem{Stoudenmire2018-3}
E.~M. Stoudenmire, {\em Learning relevant features of data with multi-scale
  tensor networks}, \href{https://doi.org/10.1088/2058-9565/aaba1a} {Quantum
  Sci. Technol. {\bf 3},  034003  (2018)}.

\bibitem{Grant2018-4}
E. Grant, M. Benedetti, S. Cao, A. Hallam, J. Lockhart, V. Stojevic, A.~G.
  Green, and S. Severini, {\em Hierarchical quantum classifiers},
  \href{https://doi.org/10.1038/s41534-018-0116-9} {npj Quantum Inf. {\bf 4},
  65  (2018)}.

\bibitem{Liu2019-21}
D. Liu, S.-J. Ran, P. Wittek, C. Peng, R.~B. Garc{\'{\i}}a, G. Su, and M.
  Lewenstein, {\em Machine learning by unitary tensor network of hierarchical
  tree structure}, \href{https://doi.org/10.1088/1367-2630/ab31ef} {New J.
  Phys. {\bf 21},  073059  (2019)}.

\bibitem{Huggins2019-4}
W. Huggins, P. Patil, B. Mitchell, K.~B. Whaley, and E.~M. Stoudenmire, {\em
  Towards quantum machine learning with tensor networks},
  \href{https://doi.org/10.1088/2058-9565/aaea94} {Quantum Sci. Technol. {\bf
  4},  024001  (2019)}.

\bibitem{Efthymiou2019_06}
S. Efthymiou, J. Hidary, and S. Leichenauer, {\em TensorNetwork for machine
  learning}, \href{https://doi.org/10.48550/arXiv.1906.06329} {arXiv:1906.06329
   (2019)}.

\bibitem{Glasser2020-8}
I. Glasser, N. Pancotti, and J.~I. Cirac, {\em From probabilistic graphical
  models to generalized tensor networks for supervised learning},
  \href{https://doi.org/10.1109/ACCESS.2020.2986279} {IEEE Access {\bf 8},
  68169  (2020)}.

\bibitem{Selvan2020_04}
R. Selvan and E.~B. Dam, {\em Tensor networks for medical image
  classification}, \href{https://doi.org/10.48550/arXiv.2004.10076}
  {International Conference on Medical Imaging with Deep Learning  721
  (2020)}.

\bibitem{Chen2020_11}
S.~Y.-C. Chen, C.-M. Huang, C.-W. Hsing, and Y.-J. Kao, {\em Hybrid
  quantum-classical classifier based on tensor network and variational quantum
  circuit}, \href{http://arxiv.org/abs/2011.14651} {arXiv:2011.14651  (2020)}.

\bibitem{Liu2021-7}
Y. Liu, W.-J. Li, X. Zhang, M. Lewenstein, G. Su, and S.-J. Ran, {\em
  Entanglement-based feature extraction by tensor network machine learning},
  \href{https://doi.org/10.3389/fams.2021.716044} {Front. Appl. Math. Stat.
  {\bf 7},  716044  (2021)}.

\bibitem{Araz2021-2021}
J.~Y. Araz and M. Spannowsky, {\em Quantum-inspired event reconstruction with
  tensor networks: matrix product states},
  \href{https://doi.org/10.1007/JHEP08(2021)112} {J. High Energy Phys. {\bf
  2021},  112  (2021)}.

\bibitem{Cheng2021-103}
S. Cheng, L. Wang, and P. Zhang, {\em Supervised learning with projected
  entangled pair states}, \href{https://doi.org/10.1103/PhysRevB.103.125117}
  {Phys. Rev. B {\bf 103},  125117  (2021)}.

\bibitem{Kong2021_01}
F. Kong, X. yang Liu, and R. Henao, {\em Quantum tensor network in machine
  learning: An application to tiny object classification},
  \href{http://arxiv.org/abs/2101.03154} {arXiv:2101.03154  (2021)}.

\bibitem{Dborin2022-7}
J. Dborin, F. Barratt, V. Wimalaweera, L. Wright, and A.~G. Green, {\em Matrix
  product state pre-training for quantum machine learning},
  \href{https://doi.org/10.1088/2058-9565/ac7073} {Quantum Science and
  Technology {\bf 7},  035014  (2022)}.

\bibitem{Convy2022-3a}
I. Convy, W. Huggins, H. Liao, and K.~B. Whaley, {\em Mutual information
  scaling for tensor network machine learning},
  \href{https://doi.org/10.1088/2632-2153/ac44a9} {Mach. Learn.: Sci. Technol.
  {\bf 3},  015017  (2022)}.

\bibitem{Convy2022-3b}
I. Convy and K.~B. Whaley, {\em Interaction decompositions for tensor network
  regression}, \href{https://doi.org/10.1088/2632-2153/aca271} {Mach. Learn.:
  Sci. Technol. {\bf 3},  045027  (2022)}.

\bibitem{Dilip2022-4}
R. Dilip, Y.-J. Liu, A. Smith, and F. Pollmann, {\em Data compression for
  quantum machine learning},
  \href{https://doi.org/10.1103/physrevresearch.4.043007} {Phys. Rev. Res. {\bf
  4},    (2022)}.

\bibitem{Dymarsky2022-4}
A. Dymarsky and K. Pavlenko, {\em Tensor network to learn the wave function of
  data}, \href{https://doi.org/10.1103/PhysRevResearch.4.043111} {Phys. Rev.
  Res. {\bf 4},  043111  (2022)}.

\bibitem{Strashko2022_08}
A. Strashko and E.~M. Stoudenmire, {\em Generalization and overfitting in
  matrix product state machine learning architectures},
  \href{http://arxiv.org/abs/2208.04372} {arXiv:2208.04372  (2022)}.

\bibitem{Guala2023-13}
D. Guala, S. Zhang, E. Cruz, C.~A. Riofr{\'i}o, J. Klepsch, and J.~M. Arrazola,
  {\em Practical overview of image classification with tensor-network quantum
  circuits}, \href{https://doi.org/10.1038/s41598-023-30258-y} {Sci. Rep. {\bf
  13},  4427  (2023)}.

\bibitem{Han2018-8}
Z.-Y. Han, J. Wang, H. Fan, L. Wang, and P. Zhang, {\em Unsupervised generative
  modeling using matrix product states},
  \href{https://doi.org/10.1103/PhysRevX.8.031012} {Phys. Rev. X {\bf 8},
  031012  (2018)}.

\bibitem{Cheng2019-99}
S. Cheng, L. Wang, T. Xiang, and P. Zhang, {\em Tree tensor networks for
  generative modeling}, \href{https://doi.org/10.1103/PhysRevB.99.155131}
  {Phys. Rev. B {\bf 99},  155131  (2019)}.

\bibitem{Sun2020-101}
Z.-Z. Sun, C. Peng, D. Liu, S.-J. Ran, and G. Su, {\em Generative tensor
  network classification model for supervised machine learning},
  \href{https://doi.org/10.1103/PhysRevB.101.075135} {Phys. Rev. B {\bf 101},
  075135  (2020)}.

\bibitem{Bai2022-39}
S.-C. Bai, Y.-C. Tang, and S.-J. Ran, {\em Unsupervised recognition of
  informative features via tensor network machine learning and quantum
  entanglement variations},
  \href{https://doi.org/10.1088/0256-307X/39/10/100701} {Chinese Phys. Lett.
  {\bf 39},  100701  (2022)}.

\bibitem{Shi2022-105}
X. Shi, Y. Shang, and C. Guo, {\em Clustering using matrix product states},
  \href{https://doi.org/10.1103/PhysRevA.105.052424} {Phys. Rev. A {\bf 105},
  052424  (2022)}.

\bibitem{Fernandez2022-12}
Y. N\'u\~nez Fern\'andez, M. Jeannin, P.~T. Dumitrescu, T. Kloss, J. Kaye, O.
  Parcollet, and X. Waintal, {\em Learning Feynman diagrams with tensor
  trains}, \href{https://doi.org/10.1103/PhysRevX.12.041018} {Phys. Rev. X {\bf
  12},  041018  (2022)}.

\bibitem{Vieijra2022_02}
T. Vieijra, L. Vanderstraeten, and F. Verstraete, {\em Generative modeling with
  projected entangled-pair states}, \href{http://arxiv.org/abs/2202.08177}
  {arXiv:2202.08177  (2022)}.

\bibitem{Lidiak2022_07}
A. Lidiak, C. Jameson, Z. Qin, G. Tang, M.~B. Wakin, Z. Zhu, and Z. Gong, {\em
  Quantum state tomography with tensor train cross approximation},
  \href{http://arxiv.org/abs/2207.06397} {arXiv:2207.06397  (2022)}.

\bibitem{Liu2023-107}
J. Liu, S. Li, J. Zhang, and P. Zhang, {\em Tensor networks for unsupervised
  machine learning}, \href{https://doi.org/10.1103/PhysRevE.107.L012103} {Phys.
  Rev. E {\bf 107},  L012103  (2023)}.

\bibitem{Shi2023_02}
X. Shi and Y. Shang, {\em Density peak clustering using tensor network},
  \href{http://arxiv.org/abs/2302.00192} {arXiv:2302.00192  (2023)}.

\bibitem{Chen2022_05}
H. Chen and T. Barthel, {\em Tensor network states with low-rank tensors},
  \href{http://arxiv.org/abs/2205.15296} {arXiv:2205.15296  (2022)}.

\bibitem{Hitchcock1927-6}
F.~L. Hitchcock, {\em The expression of a tensor or a polyadic as a sum of
  products}, \href{https://doi.org/10.1002/sapm192761164} {J. Math. Phys. {\bf
  6},  164  (1927)}.

\bibitem{Carroll1970-35}
J.~D. Carroll and J.-J. Chang, {\em Analysis of individual differences in
  multidimensional scaling via an n-way generalization of ``Eckart-Young''
  decomposition}, \href{https://doi.org/10.1007/BF02310791} {Psychometrika {\bf
  35},  283  (1970)}.

\bibitem{Harshman1970-16}
R. Harshman, {\em Foundations of the PARAFAC procedure: Models and conditions
  for an ``explanatory'' multi-modal factor analysis}, {UCLA working papers in
  phonetics {\bf 16},  1  (1970)}.

\bibitem{Kolda2009-51}
T.~G. Kolda and B.~W. Bader, {\em Tensor decompositions and applications},
  \href{https://doi.org/10.1137/07070111X} {SIAM Rev. {\bf 51},  455  (2009)}.

\bibitem{Nishino1996-65}
T. Nishino and K. Okunishi, {\em Corner transfer matrix renormalization group
  method}, \href{https://doi.org/10.1143/JPSJ.65.891} {J. Phys. Soc. Jpn. {\bf
  65},  891  (1996)}.

\bibitem{Orus2009_05}
R. Or\'{u}s and G. Vidal, {\em Simulation of two-dimensional quantum systems on
  an infinite lattice revisited: Corner transfer matrix for tensor
  contraction}, \href{https://doi.org/10.1103/PhysRevB.80.094403} {Phys. Rev. B
  {\bf 80},  094403  (2009)}.

\bibitem{Barthel2022-112}
T. Barthel, J. Lu, and G. Friesecke, {\em On the closedness and geometry of
  tensor network state sets}, \href{https://doi.org/10.1007/s11005-022-01552-z}
  {Lett. Math. Phys. {\bf 112},  72  (2022)}.

\bibitem{Miao2021_08}
Q. Miao and T. Barthel, {\em Quantum-classical eigensolver using multiscale
  entanglement renormalization},
  \href{https://doi.org/10.1103/PhysRevResearch.5.033141} {Phys. Rev. Research
  {\bf 5},  033141  (2023)}.

\bibitem{Bengio1993-3}
Y. Bengio, P. Frasconi, and P. Simard, {\em The problem of learning long-term
  dependencies in recurrent networks},
  \href{https://doi.org/10.1109/ICNN.1993.298725} {IEEE International
  Conference on Neural Networks {\bf 3},  1183  (1993)}.

\bibitem{Hochreiter1998-06}
S. Hochreiter, {\em The vanishing gradient problem during learning recurrent
  neural nets and problem solutions},
  \href{https://doi.org/10.1142/S0218488598000094} {Int. J. Uncertain.
  Fuzziness Knowl.-Based Syst. {\bf 06},  107  (1998)}.

\bibitem{Fukumizu2000-13}
K. Fukumizu and S. Amari, {\em Local minima and plateaus in hierarchical
  structures of multilayer perceptrons},
  \href{https://doi.org/10.1016/S0893-6080(00)00009-5} {Neural Netw. {\bf 13},
  317  (2000)}.

\bibitem{Dauphin2014-2}
Y.~N. Dauphin, R. Pascanu, C. Gulcehre, K. Cho, S. Ganguli, and Y. Bengio, {\em
  Identifying and attacking the saddle point problem in high-dimensional
  non-convex optimization}, \href{https://doi.org/10.48550/arXiv.1406.2572}
  {Advances in Neural Information Processing Systems {\bf 27},  2933–2941
  (2014)}.

\bibitem{Shalev2017-70}
S. Shalev-Shwartz, O. Shamir, and S. Shammah, {\em Failures of gradient-based
  deep learning}, \href{https://doi.org/10.48550/arXiv.1703.07950} {Proc.
  Machine Learning Research {\bf 70},  3067  (2017)}.

\bibitem{Barthel2023_03}
T. Barthel and Q. Miao, {\em Absence of barren plateaus and scaling of
  gradients in the energy optimization of isometric tensor network states},
  \href{http://arxiv.org/abs/2304.00161, accepted in Commun.\ Math.\ Phys.}
  {arXiv:2304.00161, accepted in Commun.\ Math.\ Phys.  (2023)}.

\bibitem{Miao2024-109}
Q. Miao and T. Barthel, {\em Isometric tensor network optimization for
  extensive Hamiltonians is free of barren plateaus},
  \href{https://doi.org/10.1103/PhysRevA.109.L050402} {Phys. Rev. A {\bf 109},
  L050402  (2024)}.

\bibitem{Liu2022-129}
Z. Liu, L.-W. Yu, L.-M. Duan, and D.-L. Deng, {\em Presence and absence of
  barren plateaus in tensor-network based machine learning},
  \href{https://doi.org/10.1103/PhysRevLett.129.270501} {Phys. Rev. Lett. {\bf
  129},  270501  (2022)}.

\bibitem{Kolbeinsson2021-15}
A. Kolbeinsson, J. Kossaifi, Y. Panagakis, A. Bulat, A. Anandkumar, I.
  Tzoulaki, and P.~M. Matthews, {\em Tensor dropout for robust learning},
  \href{https://doi.org/10.1109/jstsp.2021.3064182} {IEEE J. Select. Top.
  Signal Process. {\bf 15},  630  (2021)}.

\bibitem{Chen2016_08}
T. Chen and C. Guestrin, {\em {XGBoost}: A scalable tree boosting system},
  \href{https://doi.org/10.1145/2939672.2939785} {Proceedings of the 22nd {ACM}
  {SIGKDD} International Conference on Knowledge Discovery and Data Mining
  (2016)}.

\bibitem{Xiao2017_08}
H. Xiao, K. Rasul, and R. Vollgraf, {\em Fashion-MNIST: a novel image dataset
  for benchmarking machine learning algorithms},
  \href{https://doi.org/10.48550/ARXIV.1708.07747} {arXiv:1708.07747  (2017)}.

\bibitem{Krizhevsky2017-60}
A. Krizhevsky, I. Sutskever, and G.~E. Hinton, {\em {ImageNet} classification
  with deep convolutional neural networks},
  \href{https://doi.org/10.1145/3065386} {Commun. {ACM} {\bf 60},  84  (2017)}.

\bibitem{Mohri2018}
M. Mohri, A. Rostamizadeh, and A. Talwalkar, {\em Foundations of Machine
  Learning}, {\em Adaptive Computation and Machine Learning}, 2nd ed. (MIT
  Press, Cambridge, MA, 2018).

\bibitem{Murphy2013}
K.~P. Murphy, {\em Machine Learning: A Probabilistic Perspective} (MIT Press,
  Cambridge, MA, 2013).

\bibitem{Lecun1998-86}
Y. Lecun, L. Bottou, Y. Bengio, and P. Haffner, {\em Gradient-based learning
  applied to document recognition}, \href{https://doi.org/10.1109/5.726791}
  {Proc. IEEE {\bf 86},  2278  (1998)}.

\bibitem{Kingma2014_12}
D.~P. Kingma and J. Ba, {\em Adam: A method for stochastic optimization},
  \href{https://doi.org/10.48550/arxiv.1412.6980} {International Conference on
  Learning Representations  (2015)}.

\bibitem{Abadi2015}
M. Abadi, A. Agarwal, P. Barham, E. Brevdo, Z. Chen, C. Citro, G.~S. Corrado,
  A. Davis, J. Dean, M. Devin, {\it et~al.}, {TensorFlow}: Large-scale machine
  learning on heterogeneous systems, 2015.

\bibitem{Loshchilov2017_11}
I. Loshchilov and F. Hutter, {\em Decoupled weight decay regularization},
  \href{https://doi.org/10.48550/arXiv.1711.05101} {International Conference on
  Learning Representations  (2019)}.

\bibitem{Reddi2019_04}
S.~J. Reddi, S. Kale, and S. Kumar, {\em On the convergence of Adam and
  beyond}, \href{https://doi.org/10.48550/arXiv.1904.09237} {International
  Conference on Learning Representations  (2018)}.

\bibitem{Srivastava2014-15}
N. Srivastava, G. Hinton, A. Krizhevsky, I. Sutskever, and R. Salakhutdinov,
  {\em Dropout: a simple way to prevent neural networks from overfitting}, {J.
  Mach. Learn. Res. {\bf 15},  1929  (2014)}.

\bibitem{Bini1979-8}
D. Bini, M. Capovani, F. Romani, and G. Lotti, {\em $O(n^{2.7799})$ complexity
  for $n\times n$ approximate matrix multiplication},
  \href{https://doi.org/10.1016/0020-0190(79)90113-3} {Inform. Process. Lett.
  {\bf 8},  234  (1979)}.

\bibitem{DeSilva2008-30}
V. de~Silva and L.-H. Lim, {\em Tensor rank and the ill-posedness of the best
  low-rank approximation problem}, \href{https://doi.org/10.1137/06066518X}
  {SIAM J. Matrix Anal. Appl. {\bf 30},  1084  (2008)}.

\bibitem{Martyn2020_07}
J. Martyn, G. Vidal, C. Roberts, and S. Leichenauer, {\em Entanglement and
  Tensor Networks for Supervised Image Classification},
  \href{http://arxiv.org/abs/2007.06082} {arXiv:2007.06082  (2020)}.

\end{thebibliography}
\end{document}